\pdfoutput=1

\documentclass[11pt]{article}

\usepackage{acl}

\usepackage{times}
\usepackage{multirow}
\usepackage{xspace}
\usepackage{latexsym}
\usepackage{graphicx}
\usepackage{hyperref}
\usepackage{array}
\usepackage{color, colortbl}
\usepackage{xcolor}
\usepackage[inline]{enumitem}
\usepackage{moresize}
\usepackage{booktabs}

\usepackage[T1]{fontenc}

\usepackage[utf8]{inputenc}

\usepackage{microtype}

\def\founta{\textsc{Founta}\xspace}
\def\davidson{\textsc{Davidson}\xspace}
\def\hatexplain{\textsc{HateXplain}\xspace}
\def\volkova{\textsc{Volkova}\xspace}
\def\groenwold{\textsc{Groenwold}\xspace}
\def\bert{\texttt{BERT}\xspace}
\def\lime{{\fontfamily{lmr}\selectfont LIME}\xspace}
\long\def\eat#1{}

\def\figlabel#1{\label{fig:#1}\label{p:#1}}

\definecolor{olivegreen}{RGB}{0,128,0}
\definecolor{orange}{RGB}{225,128,0}
\definecolor{brickred}{RGB}{230,0,0}

\newcounter{notecounter}
\newcommand{\enotesoff}{\long\gdef\enote##1##2{}}
\newcommand{\enoteson}{\long\gdef\enote##1##2{{
			\stepcounter{notecounter}
			{\large\textbf{ \hspace{1cm}\arabic{notecounter} $<<<$ ##1: ##2 $>>>$\hspace{1cm}}}}}}
\enoteson
\enotesoff

\title{Analyzing Hate Speech Data along Racial, Gender and Intersectional Axes}

\author{Antonis Maronikolakis\Thanks{ Equal contribution.} \and 
	Philip Baader$^*$  \and
	Hinrich Sch\"{u}tze \\
	Center for Information and Language Processing (CIS), LMU Munich, Germany\\
	{\tt antmarakis@cis.lmu.de}}

\begin{document}
\maketitle

\begin{abstract}

\textbf{\textcolor{red!100}{Warning:}} \textit{This work contains strong and offensive language, sometimes uncensored.}

To tackle the rising phenomenon of hate speech, efforts have been made towards data curation and analysis. When it comes to analysis of bias, previous work has focused predominantly on race. In our work, we further investigate bias in hate speech datasets along racial, gender and intersectional axes.
We identify strong bias against African American English (AAE), masculine and AAE+Masculine tweets, which are annotated as disproportionately more hateful and offensive than from other demographics. We provide evidence that \bert-based models propagate this bias and show that balancing the training data for these protected attributes can lead to fairer models with regards to gender, but not race.

\end{abstract}

\section{Introduction}

\textbf{Hate Speech.} To tackle the phenomenon of online hate speech, efforts have been made to curate datasets \cite{davidson,misogyny_annotated_data,social_bias_frames}. Since datasets in this domain are dealing with sensitive topics, it is of upmost importance that biases are kept to a (realistic) minimum and that data is thoroughly analyzed before use \cite{racial_bias_in_hatespeech,hate_speech_data_analysis}. In our work, we are contributing to this analysis by uncovering biases along the racial, gender and intersectional axes.

\textbf{Racial\footnote{While the correlation of race and African American English (AAE) is complicated \cite{anderson}, in our work we consider AAE as a proxy for race, since it is a dialect overwhelmingly used by African Americans \cite{spears1,spears2}.}, Gender and Intersectional Biases.} During data collection, biases can be introduced due to--among other reasons--lack of annotator training or divergence between annotators and user demographics. For example, oftentimes the majority of annotators are white or male \cite{social_bias_frames,founta}. An annotator not in the `in-group' may hold (un)conscious biases based on misconceptions about `in-group' speech which may affect their perception of speech from certain communities \cite{factors_of_offensiveness}, leading to incorrect annotations when it comes to dialects the annotators are not familiar with. A salient example of this is annotators conflating African American English with hateful language \cite{aae_bias_hatespeech}.

\begin{figure}[!t]
	\centering
	\includegraphics[scale=0.235]{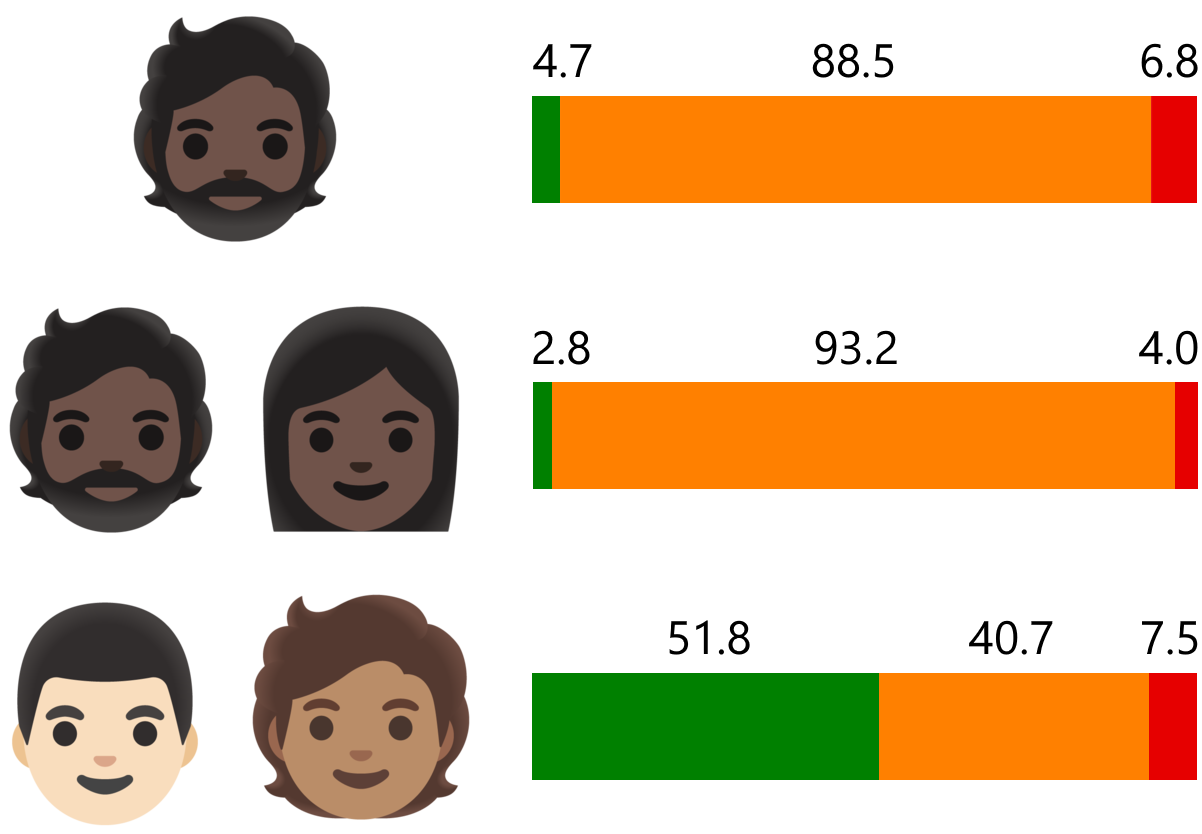}
	\caption{Distributions of label annotations on \davidson (\textcolor{olivegreen}{neutral}, \textcolor{orange}{offensive}, \textcolor{brickred}{hateful}) for AAE+Masculine, AAE and SAE (top-to-bottom). AAE has a higher ratio of offensive examples than SAE, while AAE+Masculine is both highly offensive and hateful.}
	\figlabel{pitch}
\end{figure}

Intersectionality \cite{kimberle} is a framework for examining how different forms of inequality (for example, racial or gender inequalities) intersect with and reinforce each other. These new social dynamics need to be analyzed both separately and as a whole in order to address challenges faced by the examined communities. For example, a black woman does not face inequality based only on race or only on gender: she faces inequality because of both these characteristics, separately and in conjunction. In this work, we are analyzing not only the racial or gender inequalities in hate speech datasets, but their intersectionality as well.

With research in the area of hate speech, the NLP community aims at protecting target groups and fostering a safer online environment. In this sensitive area, it is pivotal that datasets and models are analyzed extensively to ensure the biases we are protecting affected communities from do not appear in the data itself, causing further marginalization (for example, by removing AAE speech disproportionately more often).

\textbf{Contributions.} In summary, we  \begin{enumerate*}[label={\textbf{(\roman{*})}}]
	\item investigate racial, gender and intersectional bias in three hate speech datasets, \citet{founta,davidson,hatexplain},
	\item examine classifier predictions on existing, general-purpose African/Standard American English (AAE/SAE) and gendered tweets,
	\item identify model bias against AAE, masculine and AAE+Masculine (labeled as both AAE and masculine) tweets,
	\item show that balancing training data for gender leads to fairer models
\end{enumerate*}.

\section{Related Work}

Hate speech research has focused on dataset curation \cite{davidson,founta,social_bias_frames,misogyny_annotated_data,incivility_in_news,hate_towards_political_opponent} and dataset analysis \cite{hate_speech_data_analysis,hatespeech_biased_datasets,hate_speech_cross_dataset}. In our work, we further analyze datasets to uncover latent biases.

It has been shown that data reflects social bias inherent in annotator pools 
\cite{are_you_racist,annotator_bias_demographic,racial_bias_in_hatespeech,racial_bias_in_data}.
Work has been conducted to identify bias against AAE \cite{aae_bias_hatespeech,hatespeech_debias_words,adversarial_training_aae_hatespeech} and gender \cite{excell-al-moubayed-2021-towards}.

Research has also been conducted in identifying disparities in performance across social groups, with machine learning algorithms underperforming for certain groups \cite{tatman-2017-gender,pmlr-v81-buolamwini18a,rudinger-etal-2018-gender}.

\citet{intersectional_bias} investigated whether bias along the intersectional axis exists in \citet{founta}. While \citet{intersectional_bias} focused on bias within a single dataset, in our work we generalize to multiple hate speech datasets. We also examine classifier behavior and methods to mitigate this bias.

Research from a sociolinguistic perspective has shown that genders exhibit differences in online text \cite{gender_online_lexicon_diff} as well as general speech \cite{language_and_gender}. In \citet{gender_bamman} and \citet{gender_bergsma}, gender classifiers for English tweets were developed with accuracy of 88\% and 85\% respectively. In our work, we develop a gender classifier of tweets as well, focusing on precision over recall, leading to a smaller but more accurate sample of gendered data.

\begin{table}[!t]
	\centering
	\begin{tabular}{cccc} \toprule
		Dataset & Neutral & Offensive & Hateful \\ \midrule
		\davidson & 0.92 & 0.85 & 0.53 \\
		\founta & 0.85 & 0.79 & 0.47 \\
		\hatexplain & 0.69 & 0.55 & 0.70 \\ \bottomrule
	\end{tabular}
	\caption{F1-score of \bert for each label, evaluated on \davidson, \founta and \hatexplain.}
	\label{table:hate_classifier}
\end{table}

\section{Datasets}

Five English datasets were used: three hate speech datasets (\davidson, \founta and \hatexplain), one dataset of tweets labeled for race (\groenwold) and one for gender (\volkova). We adopt the definitions of \citet{davidson} for hate speech (defined as speech that contains expressions of hatred towards a group or individual on the basis of protected attributes like ethnicity, gender, race and sexual orientation) and offensive speech (speech that contains offensive language but is not hateful).

\textbf{\davidson.} In \citet{davidson}, a hate speech dataset of tweets was collected, labeled for neutral, offensive and hateful language.

\textbf{\founta.} In \citet{founta} a crowdsourced dataset of tweets was presented, labeled for normal, abusive and hateful language. To unify definitions, we rename normal to neutral language and abusive to offensive language.

\textbf{\hatexplain.} \citet{hatexplain} presented a dataset from Twitter and Gab\footnote{Gab is a social platform that has been known to host far-right groups and rhetoric.} passages. It has been labeled for normal (neutral), offensive and hateful language.

\textbf{\groenwold.} In \citet{groenwold} a dataset of African American English and Standard American English tweets was introduced. The AAE tweets come from \cite{blodgett} and the SAE are direct translations of those tweets provided by annotators.

\textbf{\volkova.} \citet{volkova} presented a dataset of 800k English tweets from users with an associated gender (feminine/masculine).

\begin{table*}[!t]
	\small
	\centering
	\begin{tabular}{ccccccccc} \toprule
		Dataset & Masc. & Fem. & SAE & AAE & SAE+Masc. & SAE+Fem. & AAE+Masc. & AAE+Fem. \\ \midrule
		\davidson & 2716 & 2338 & 3534 & 8099 & 1279 & 1240 & 3157 & 1172 \\
		\founta & 26307 & 13615 & 43330 & 4177 & 13486 & 13257 & 971 & 787 \\
		\hatexplain & 4509 & 1103 & 10368 & 1103 & 4145 & 2376 & 250 & 240 \\
		\groenwold$_{AAE}$ & 586 & 613 & 0 & 1995 & 0 & 0 & 587 & 612 \\
		\groenwold$_{SAE}$ & 587 & 601 & 1980 & 0 & 587 & 601 & 0 & 0 \\
		\volkova & 41164 & 58836 & 37874 & 3755 & 16243 & 21631 & 1843 & 1912 \\ \bottomrule
	\end{tabular}
	\caption{Protected attribute statistics for \davidson, \founta, \hatexplain, \groenwold and \volkova.}
	\label{table:inter_statistics}
\end{table*}

\section{Experimental Setup}
\label{setup}

\textbf{AAE Classifier.} To classify tweets as AAE or SAE, we used the \citet{blodgett} classifier. We took into consideration tweets with a confidence score over 0.5, which can be interpreted as a straightforward classifier of AAE/SAE (whichever class has the highest score is returned).

\textbf{Gender Classifier.} To classify tweets as masculine or feminine, we finetuned \bert-base\footnote{\url{https://huggingface.co/bert-base-cased}} on \citet{volkova}, which includes gender information as self-reported from authors. We split the dataset into train/dev/test (50K/25K/25K) and employed a confidence score of 0.8 as the threshold for assigning gender to a tweet. For the tweets with a confidence over the given threshold, precision was 78.4\% when classifying tweets as `masculine' and 79.5\% when classifying tweets as `feminine'.

\textbf{Hate Speech Classifiers.} For each of the three hate speech datasets we finetuned \bert-base. We split each dataset into 80:10:10 (train:dev:test) sets, used a max sequence length of 256 and trained for 3 epochs, keeping the rest of the hyperparameters the same.
Performance for the development set is shown in Table \ref{table:hate_classifier}\footnote{Performance on the test set is similar, omitted for brevity.}. In \davidson and \founta, \bert performs well for neutral and offensive examples, performance drops for hateful content. In \hatexplain, \bert overall performs worse, with slightly better performance for neutral and hateful examples over offensive ones.

\textbf{Intersectionality.} For our analysis, we classified tweets from all datasets for gender and race.

\section{Intersectionality Statistics}

In Table \ref{table:inter_statistics}, we present statistics for gender, race and their intersection as found in the three examined hate speech datasets as well as in \groenwold and \volkova.\footnote{Race/gender for the hate speech datasets, gender for \groenwold and race for \volkova have been computed as described in Section \ref{setup}.} We show that no dataset is balanced between AAE and SAE. In \founta and \hatexplain, AAE tweets make up approximately 1/10th of the data. In \davidson, we see stronger representation of AAE, with the AAE tweets being almost twice as many as the SAE tweets. \davidson is also balanced for gender. The other hate speech datasets, while still not balanced, are more balanced for gender than they are for race. \founta has twice as many masculine than feminine tweets and \hatexplain has four times as many.

In Table \ref{table:inter_class_stats}, we present a breakdown of protected attributes per class (neutral/offensive/hateful) for \davidson, \founta and \hatexplain. A main takeaway for \davidson and \founta is the imbalance of AAE versus SAE. In SAE, the neutral class makes up 52\% of the data for \davidson and 81\% for \founta, while the respective numbers for AAE are 3\% for \davidson and 13\% for \founta.

In \hatexplain, AAE and SAE are more balanced, but there is instead imbalance between genders. For masculine and feminine speech, passages are neutral at rates of 43\% and 61\% respectively. In \davidson, SAE+Feminine speech is viewed as more offensive than SAE+Masculine (48\% vs. 19\%), while in \hatexplain, SAE+Masculine is more hateful than SAE+Feminine (34\% vs. 16\%). Finally, when comparing genders in AAE speech, we see that while AAE+Feminine contains a larger percentage of offensive tweets (for example, in \founta, 69\% vs. 54\% and in \hatexplain, 50\% vs. 21\%), AAE+Masculine contains disproportionately more hateful speech (in \davidson, 7\% vs. 5\%, in \founta, 28\% vs. 9\% and in \hatexplain, 19\% vs. 6\%).

Overall, AAE and masculine speech is annotated as more offensive and hateful than SAE and feminine speech. Further analyzing AAE, AAE+Masculine is viewed as more hateful than AAE+Feminine.

\begin{table*}[!t]
	\small
	\centering
	\addtolength{\tabcolsep}{-3.75pt}
	\begin{tabular}{c|ccc|ccc|ccc|ccc|ccc|ccc|ccc|ccc} \toprule
		\multirow{2}{*}{{\small Dataset}} & \multicolumn{3}{c|}{{\small Masc.}} & \multicolumn{3}{c|}{{\small Fem.}} & \multicolumn{3}{c|}{{\small SAE}} & \multicolumn{3}{c|}{{\small AAE}} & \multicolumn{3}{c|}{{\small SAE+Masc.}} & \multicolumn{3}{c|}{{\small SAE+Fem.}} & \multicolumn{3}{c|}{{\small AAE+Masc.}} & \multicolumn{3}{c}{{\small AAE+Fem.}} \\
		& \textsf{N} & \textsf{O} & \textsf{H} & \textsf{N} & \textsf{O} & \textsf{H} & \textsf{N} & \textsf{O} & \textsf{H} & \textsf{N} & \textsf{O} & \textsf{H} & \textsf{N} & \textsf{O} & \textsf{H} & \textsf{N} & \textsf{O} & \textsf{H} & \textsf{N} & \textsf{O} & \textsf{H} & \textsf{N} & \textsf{O} & \textsf{H} \\ \midrule
		\multicolumn{1}{c|}{{\small Davidson}} & {\scriptsize 32.2} & {\scriptsize 61.9} & {\scriptsize 5.9} & {\scriptsize 27.7} & {\scriptsize 69.5} & {\scriptsize 2.8} & {\scriptsize 51.8} & {\scriptsize 40.7} & {\scriptsize 7.5} & {\scriptsize 2.8} & {\scriptsize 93.2} & {\scriptsize 4.0} & {\scriptsize 77.0} & {\scriptsize 19.4} & {\scriptsize 3.6} & {\scriptsize 50.0} & {\scriptsize 47.8} & {\scriptsize 2.3} & {\scriptsize 4.7} & {\scriptsize 88.5} & {\scriptsize 6.8} & {\scriptsize 6.8} & {\scriptsize 88.0} & {\scriptsize 5.2} \\
		\multicolumn{1}{c|}{{\small Founta}} & {\scriptsize 81.2} & {\scriptsize 12.3} & {\scriptsize 6.4} & {\scriptsize 71.0} & {\scriptsize 25.0} & {\scriptsize 4.0} & {\scriptsize 80.5} & {\scriptsize 14.6} & {\scriptsize 4.9} & {\scriptsize 13.2} & {\scriptsize 69.2} & {\scriptsize 17.6} & {\scriptsize 86.9} & {\scriptsize 7.6} & {\scriptsize 5.5} & {\scriptsize 86.2} & {\scriptsize 11.4} & {\scriptsize 2.4} & {\scriptsize 18.3} & {\scriptsize 53.8} & {\scriptsize 27.9} & {\scriptsize 21.8} & {\scriptsize 69.4} & {\scriptsize 8.8} \\
		\multicolumn{1}{c|}{{\small HateXplain}} & {\scriptsize 43.0} & {\scriptsize 23.7} & {\scriptsize 33.3} & {\scriptsize 60.7} & {\scriptsize 24.6} & {\scriptsize 14.8} & {\scriptsize 38.3} & {\scriptsize 26.7} & {\scriptsize 35.0} & {\scriptsize 45.6} & {\scriptsize 39.1} & {\scriptsize 15.3} & {\scriptsize 41.6} & {\scriptsize 24.0} & {\scriptsize 34.4} & {\scriptsize 58.9} & {\scriptsize 25.1} & {\scriptsize 16.0} & {\scriptsize 59.4} & {\scriptsize 21.3} & {\scriptsize 19.4} & {\scriptsize 44.4} & {\scriptsize 50.0} & {\scriptsize 5.6} \\ \bottomrule
	\end{tabular}
	\addtolength{\tabcolsep}{3.75pt}
	\caption{Distribution of protected attribute annotations for neutral/offensive/hateful (\textsf{N}/\textsf{O}/\textsf{H}) examples.}
	\label{table:inter_class_stats}
\end{table*}

\begin{table*}[!t]
\small
\centering
\addtolength{\tabcolsep}{-3.75pt}
\begin{tabular}{c|ccc|ccc|ccc|ccc|ccc|ccc|ccc|ccc} \toprule
	\multirow{2}{*}{{\small Dataset}} & \multicolumn{3}{c|}{Masc.} & \multicolumn{3}{c|}{Fem.} & \multicolumn{3}{c|}{SAE} & \multicolumn{3}{c|}{AAE} & \multicolumn{3}{c|}{SAE+Masc.} & \multicolumn{3}{c|}{SAE+Fem.} & \multicolumn{3}{c|}{AAE+Masc.} & \multicolumn{3}{c}{AAE+Fem.} \\
	& \textsf{N} & \textsf{O} & \textsf{H} & \textsf{N} & \textsf{O} & \textsf{H} & \textsf{N} & \textsf{O} & \textsf{H} & \textsf{N} & \textsf{O} & \textsf{H} & \textsf{N} & \textsf{O} & \textsf{H} & \textsf{N} & \textsf{O} & \textsf{H} & \textsf{N} & \textsf{O} & \textsf{H} & \textsf{N} & \textsf{O} & \textsf{H} \\ \midrule
	\multicolumn{1}{c|}{Random} & {\scriptsize 33.8} & {\scriptsize 63.2} & {\scriptsize 3.0} & {\scriptsize 27.7} & {\scriptsize 71.2} & {\scriptsize 1.1} & {\scriptsize 53.1} & {\scriptsize 40.5} & {\scriptsize 6.4} & {\scriptsize 4.9} & {\scriptsize 94.1} & {\scriptsize 1.0} & {\scriptsize 77.3} & {\scriptsize 19.3} & {\scriptsize 3.4} & {\scriptsize 45.6} & {\scriptsize 53.2} & {\scriptsize 1.2} & {\scriptsize 6.4} & {\scriptsize 91.2} & {\scriptsize 2.4} & {\scriptsize 3.0} & {\scriptsize 94.3} & {\scriptsize 2.7} \\
	\multicolumn{1}{c|}{Balanced} & {\scriptsize 25.3} & {\scriptsize 71.5} & {\scriptsize 3.2} & {\scriptsize 25.4} & {\scriptsize 71.1} & {\scriptsize 3.5} & {\scriptsize 54.3} & {\scriptsize 39.2} & {\scriptsize 6.5} & {\scriptsize 4.3} & {\scriptsize 95.1} & {\scriptsize 1.6} & {\scriptsize 71.0} & {\scriptsize 22.8} & {\scriptsize 6.2} & {\scriptsize 52.3} & {\scriptsize 46.4} & {\scriptsize 2.3} & {\scriptsize 5.8} & {\scriptsize 92.1} & {\scriptsize 2.1} & {\scriptsize 6.2} & {\scriptsize 93.1} & {\scriptsize 0.7} \\ \bottomrule
\end{tabular}
\addtolength{\tabcolsep}{3.75pt}
\caption{Distribution of predictions for protected attributes on random and balanced datasets based on \davidson. The balanced set is balanced on race (equal number of AAE and SAE tweets) and gender (equal number of feminine and masculine tweets). Shown are percentages for neutral/offensive/hateful (\textsf{N}/\textsf{O}/\textsf{H}) predictions.}
\label{table:rebalanced}
\end{table*}

\begin{table}[t]
	\centering
	\small
	\begin{tabular}{cp{2.275cm}p{2.275cm}} \toprule
		Dataset & All & AAE \\ \midrule
		\davidson & n*ggerize, subhuman, bastards, border, pigfucking, feminist, wetbacks, savages, wetback, jumpers & queer, n*gros, n*ggaz, racial, shittiest, wet, savage, skinned, darky, f*gs \\ \midrule
		\founta & moron, insult, muslims, aggression, puritan, haters, arabs, coloured, ousted, pedophiles & white, killing, pathetic, n*gga, slave, n*ggas, sells, hell, children, violent \\ \midrule
		\hatexplain & towelhead, muzzrat, muscum, n*gresses, n*ggerette, n*glets, musloid, n*ggerish, n*ggery, gorilla & spic, fuck, f*ggots, gorilla, towel, sandn*gger, zhid, c*ons, rag, fowl \\ \bottomrule
	\end{tabular}
	\caption{Top 10 most contributing words for \davidson, \founta and \hatexplain as computed with \lime for hateful predictions.}
	\label{lime_res}
\end{table}

\section{Bias in BERT}

We investigate to what extent data bias is learned by \bert.
We compare our findings against a dataset balanced for race and gender, to examine whether balanced data leads to fairer models. Namely, we compare a randomly sampled with a balanced set the \davidson dataset.\footnote{\founta and \hatexplain were not considered for this study as they do not contain enough AAE examples to make confident inferences.} In the balanced set we sample the same number of AAE and SAE tweets (3000) and the same number of masculine and feminine tweets (1750). We also include 8000 neutral tweets without race or gender labels. For the randomly sampled set, for a fair comparison, we sampled the same number of tweets as the balanced set.\footnote{Experiments were conducted with the entirety of the original dataset with similar results. They are omitted for brevity.} All sampling was stratified to preserve the original label distributions. Results are shown in Table \ref{table:rebalanced}.

In the randomly sampled set, there is an imbalance both for gender and race. For gender, while masculine tweets are more hateful (3\% vs. 1\%), feminine tweets are more offensive (71\% vs. 63\%). For race, AAE is marked almost entirely as offensive (94\%), while SAE is split in neutral and offensive (53\% and 41\%). In the SAE subset of tweets, there is an imbalance between genders, with SAE+Feminine being marked disproportionately more often as offensive than SAE+Masculine (54\% vs. 19\%).

\subsection{Balanced Training}

In Table \ref{table:rebalanced}, before balancing, 34\% of masculine and 28\% of feminine tweets are marked as neutral. After balancing, these rates are both at 25\%. There is an improvement in the intersection of AAE and gender, with the distributions of AAE+Masculine and AAE+Feminine tweets converging. For SAE, SAE+Masculine and SAE+Feminine distributions converge too, although still far apart. Overall, balanced data improves fairness for gender but not for race, which potentially stems from bias in annotation.

\subsection{Interpretability with LIME}

In Table \ref{lime_res}, we show the top contributing words for offensive and hateful predictions in \davidson, \founta and \hatexplain. We see that for AAE, terms such as `n****z' and `n***a' contribute in classifying text as non-neutral even though the terms are part of African American vernacular \cite{nword_aae}, showing that this dialect is more likely to be flagged. In non-AAE speech (which includes--but is not exclusive to--SAE), we see the n-word variant with the `-er' spelling appearing more often in various forms, which is correctly picked up by the model as an offensive and hateful term. On both sets, we also see other slurs, such as `f*ggots', `moron' and `wetback' (a slur against foreigners residing in the United States, especially Mexicans) being picked up, showing the model does recognize certain slurs and offensive terms.

\section{Conclusion}

In our work, we analyze racial, gender and intersectional bias in hate speech datasets.
We show that tweets from AAE and AAE+Masculine users (as classified automatically) are labeled disproportionately more often as offensive. We further show that \bert learns this bias, flagging AAE speech as significantly more offensive than SAE. We perform interpretability analysis using \lime, showing that the inability of \bert to differentiate between variations of the n-word across dialects is a contributing factor to biased predictions.
Finally, we investigate whether training on a dataset balanced for race and gender mitigates bias. This method shows mixed results, with gender bias being mitigated more than racial bias. With our work we want to motivate further investigation in model bias not only for the usual gender and racial attributes, but also for their intersection.

\section{Bias Statement}

Research in the sphere of hate speech has produced annotated data that can be used to train classifiers to detect hate speech as found in online media. It is known that these datasets contain biases that models will potentially propagate. The \textbf{representational harm} that can be triggered is certain target groups getting their speech inadvertently censored/deleted due to existing biases that marginalize certain groups. In our work we investigate this possibility along intersectional axes (gender and race). We find that tweets written by female users are seen as disproportionately more offensive, while male users write tweets that appear more hateful.

\section{Ethical Considerations}

In our work we are dealing with data that can catalyze harm against marginalized groups.
We do not advocate for the propagation or adoption of this hateful rhetoric.
With our work we wish to motivate further analysis and documentation of sensitive data that is to be used for the training of models (for example, using templates from \citet{model_cards,data_statements}).

Further, while classifying protected attributes such as race or gender is important in analyzing and identifying bias, care should be taken for the race and gender classifiers to not be misused or abused, in order to protect the identity of users, especially those from marginalized demographics who are more vulnerable to hateful attacks and further marginalization. In our work we only predict these protected attributes for investigative purposes and do not motivate the direct application of such classifiers. Further, in our work we are using dialect (AAE) associated with African Americans as a proxy to race due to a lack of available annotated data. It should be noted that not all African Americans make use of AAE and not all AAE users are African Americans.

Finally, in our work we only focused on English and a specific set of attributes. Namely, we considered race (African American) and gender. This is a non-exhaustive list of biases and more work needs to be done for greater coverage of languages and attributes.

\section{Acknowledgments}

Antonis Maronikolakis was partly supported by the European Research Council (\#740516).

\bibliography{anthology}

\begin{thebibliography}{37}
\expandafter\ifx\csname natexlab\endcsname\relax\def\natexlab#1{#1}\fi

\bibitem[{Al~Kuwatly et~al.(2020)Al~Kuwatly, Wich, and
  Groh}]{annotator_bias_demographic}
Hala Al~Kuwatly, Maximilian Wich, and Georg Groh. 2020.
\newblock \href {https://doi.org/10.18653/v1/2020.alw-1.21} {Identifying and
  measuring annotator bias based on annotators{'} demographic characteristics}.
\newblock In \emph{Proceedings of the Fourth Workshop on Online Abuse and
  Harms}, pages 184--190, Online. Association for Computational Linguistics.

\bibitem[{Anderson(2015)}]{anderson}
Kate~T. Anderson. 2015.
\newblock Racializing language: Unpacking linguistic approaches to attitudes
  about race and speech.
\newblock In Green Lisa~J. Bloomquist, Jennifer and Sonja~L. Lanehart, editors,
  \emph{The Oxford Handbook of Innovation}, chapter~42, pages 773--785. Oxford
  University Press, Oxford.

\bibitem[{Bamman et~al.(2014)Bamman, Eisenstein, and
  Schnoebelen}]{gender_bamman}
David Bamman, Jacob Eisenstein, and Tyler Schnoebelen. 2014.
\newblock Gender identity and lexical variation in social media.
\newblock \emph{Journal of Sociolinguistics}, 18(2):135--160.

\bibitem[{Bender and Friedman(2018)}]{data_statements}
Emily~M. Bender and Batya Friedman. 2018.
\newblock \href {https://doi.org/10.1162/tacl_a_00041} {Data statements for
  natural language processing: Toward mitigating system bias and enabling
  better science}.
\newblock \emph{Transactions of the Association for Computational Linguistics},
  6:587--604.

\bibitem[{Bergsma and Van~Durme(2013)}]{gender_bergsma}
Shane Bergsma and Benjamin Van~Durme. 2013.
\newblock \href {https://aclanthology.org/P13-1070} {Using conceptual class
  attributes to characterize social media users}.
\newblock In \emph{Proceedings of the 51st Annual Meeting of the Association
  for Computational Linguistics (Volume 1: Long Papers)}, pages 710--720,
  Sofia, Bulgaria. Association for Computational Linguistics.

\bibitem[{Blodgett et~al.(2016)Blodgett, Green, and O'Connor}]{blodgett}
Su~Lin Blodgett, Lisa Green, and Brendan O'Connor. 2016.
\newblock {Demographic Dialectal Variation in Social Media: A Case Study of
  African-American English}.
\newblock In \emph{Proceedings of EMNLP}.

\bibitem[{Buolamwini and Gebru(2018)}]{pmlr-v81-buolamwini18a}
Joy Buolamwini and Timnit Gebru. 2018.
\newblock \href {https://proceedings.mlr.press/v81/buolamwini18a.html} {Gender
  shades: Intersectional accuracy disparities in commercial gender
  classification}.
\newblock In \emph{Proceedings of the 1st Conference on Fairness,
  Accountability and Transparency}, volume~81 of \emph{Proceedings of Machine
  Learning Research}, pages 77--91. PMLR.

\bibitem[{Crenshaw(1989)}]{kimberle}
Kimberle Crenshaw. 1989.
\newblock Demarginalizing the intersection of race and sex: A black feminist
  critique of antidiscrimination doctrine, feminist theory and antiracist
  politics.
\newblock \emph{The University of Chicago Legal Forum}, 140:139--167.

\bibitem[{Davidson et~al.(2019{\natexlab{a}})Davidson, Bhattacharya, and
  Weber}]{racial_bias_in_hatespeech}
Thomas Davidson, Debasmita Bhattacharya, and Ingmar Weber. 2019{\natexlab{a}}.
\newblock \href {https://doi.org/10.18653/v1/W19-3504} {Racial bias in hate
  speech and abusive language detection datasets}.
\newblock In \emph{Proceedings of the Third Workshop on Abusive Language
  Online}, pages 25--35, Florence, Italy. Association for Computational
  Linguistics.

\bibitem[{Davidson et~al.(2019{\natexlab{b}})Davidson, Bhattacharya, and
  Weber}]{racial_bias_in_data}
Thomas Davidson, Debasmita Bhattacharya, and Ingmar Weber. 2019{\natexlab{b}}.
\newblock \href {https://doi.org/10.18653/v1/W19-3504} {Racial bias in hate
  speech and abusive language detection datasets}.
\newblock In \emph{Proceedings of the Third Workshop on Abusive Language
  Online}, pages 25--35, Florence, Italy. Association for Computational
  Linguistics.

\bibitem[{Davidson et~al.(2017)Davidson, Warmsley, Macy, and Weber}]{davidson}
Thomas Davidson, Dana Warmsley, Michael Macy, and Ingmar Weber. 2017.
\newblock \href {https://aaai.org/ocs/index.php/ICWSM/ICWSM17/paper/view/15665}
  {Automated hate speech detection and the problem of offensive language}.
\newblock In \emph{International AAAI Conference on Web and Social Media}.

\bibitem[{Excell and Al~Moubayed(2021)}]{excell-al-moubayed-2021-towards}
Elizabeth Excell and Noura Al~Moubayed. 2021.
\newblock \href {https://doi.org/10.18653/v1/2021.gebnlp-1.7} {Towards equal
  gender representation in the annotations of toxic language detection}.
\newblock In \emph{Proceedings of the 3rd Workshop on Gender Bias in Natural
  Language Processing}, pages 55--65, Online. Association for Computational
  Linguistics.

\bibitem[{Founta et~al.(2018)Founta, Djouvas, Chatzakou, Leontiadis, Blackburn,
  Stringhini, Vakali, Sirivianos, and Kourtellis}]{founta}
Antigoni-Maria Founta, Constantinos Djouvas, Despoina Chatzakou, Ilias
  Leontiadis, Jeremy Blackburn, Gianluca Stringhini, Athena Vakali, Michael
  Sirivianos, and Nicolas Kourtellis. 2018.
\newblock Large scale crowdsourcing and characterization of twitter abusive
  behavior.
\newblock In \emph{11th International Conference on Web and Social Media, ICWSM
  2018}. AAAI Press.

\bibitem[{Gefen and Ridings(2005)}]{gender_online_lexicon_diff}
David Gefen and Catherine Ridings. 2005.
\newblock \href {https://doi.org/10.1145/1066149.1066156} {If you spoke as she
  does, sir, instead of the way you do: A sociolinguistics perspective of
  gender differences in virtual communities}.
\newblock \emph{DATA BASE}, 36:78--92.

\bibitem[{Grimminger and Klinger(2021)}]{hate_towards_political_opponent}
Lara Grimminger and Roman Klinger. 2021.
\newblock \href {https://aclanthology.org/2021.wassa-1.18} {Hate towards the
  political opponent: A {T}witter corpus study of the 2020 {US} elections on
  the basis of offensive speech and stance detection}.
\newblock In \emph{Proceedings of the Eleventh Workshop on Computational
  Approaches to Subjectivity, Sentiment and Social Media Analysis}, pages
  171--180, Online. Association for Computational Linguistics.

\bibitem[{Groenwold et~al.(2020)Groenwold, Ou, Parekh, Honnavalli, Levy, Mirza,
  and Wang}]{groenwold}
Sophie Groenwold, Lily Ou, Aesha Parekh, Samhita Honnavalli, Sharon Levy, Diba
  Mirza, and William~Yang Wang. 2020.
\newblock \href {https://www.aclweb.org/anthology/2020.emnlp-main.473}
  {Investigating {A}frican-{A}merican {V}ernacular {E}nglish in
  transformer-based text generation}.
\newblock In \emph{Proceedings of EMNLP}.

\bibitem[{Guest et~al.(2021)Guest, Vidgen, Mittos, Sastry, Tyson, and
  Margetts}]{misogyny_annotated_data}
Ella Guest, Bertie Vidgen, Alexandros Mittos, Nishanth Sastry, Gareth Tyson,
  and Helen Margetts. 2021.
\newblock \href {https://www.aclweb.org/anthology/2021.eacl-main.114} {An
  expert annotated dataset for the detection of online misogyny}.
\newblock In \emph{Proceedings of the 16th Conference of the European Chapter
  of the Association for Computational Linguistics: Main Volume}, pages
  1336--1350, Online. Association for Computational Linguistics.

\bibitem[{Hede et~al.(2021)Hede, Agarwal, Lu, Mutz, and
  Nenkova}]{incivility_in_news}
Anushree Hede, Oshin Agarwal, Linda Lu, Diana~C. Mutz, and Ani Nenkova. 2021.
\newblock \href {https://aclanthology.org/2021.eacl-main.225} {From toxicity in
  online comments to incivility in {A}merican news: Proceed with caution}.
\newblock In \emph{Proceedings of the 16th Conference of the European Chapter
  of the Association for Computational Linguistics: Main Volume}, pages
  2620--2630, Online. Association for Computational Linguistics.

\bibitem[{Kim et~al.(2020)Kim, Ortiz, Nam, Santiago, and
  Datta}]{intersectional_bias}
Jae{-}Yeon Kim, Carlos Ortiz, Sarah Nam, Sarah Santiago, and Vivek Datta. 2020.
\newblock \href {http://arxiv.org/abs/2005.05921} {Intersectional bias in hate
  speech and abusive language datasets}.
\newblock \emph{CoRR}, abs/2005.05921.

\bibitem[{Madukwe et~al.(2020)Madukwe, Gao, and
  Xue}]{hate_speech_data_analysis}
Kosisochukwu Madukwe, Xiaoying Gao, and Bing Xue. 2020.
\newblock \href {https://doi.org/10.18653/v1/2020.alw-1.18} {In data we trust:
  A critical analysis of hate speech detection datasets}.
\newblock In \emph{Proceedings of the Fourth Workshop on Online Abuse and
  Harms}, pages 150--161, Online. Association for Computational Linguistics.

\bibitem[{Mathew et~al.(2021)Mathew, Saha, Yimam, Biemann, Goyal, and
  Mukherjee}]{hatexplain}
Binny Mathew, Punyajoy Saha, Seid~Muhie Yimam, Chris Biemann, Pawan Goyal, and
  Animesh Mukherjee. 2021.
\newblock \href {https://ojs.aaai.org/index.php/AAAI/article/view/17745}
  {Hatexplain: A benchmark dataset for explainable hate speech detection}.
\newblock \emph{Proceedings of the AAAI Conference on Artificial Intelligence},
  35(17):14867--14875.

\bibitem[{Mitchell et~al.(2019)Mitchell, Wu, Zaldivar, Barnes, Vasserman,
  Hutchinson, Spitzer, Raji, and Gebru}]{model_cards}
Margaret Mitchell, Simone Wu, Andrew Zaldivar, Parker Barnes, Lucy Vasserman,
  Ben Hutchinson, Elena Spitzer, Inioluwa~Deborah Raji, and Timnit Gebru. 2019.
\newblock \href {https://doi.org/10.1145/3287560.3287596} {Model cards for
  model reporting}.
\newblock In \emph{Proceedings of the Conference on Fairness, Accountability,
  and Transparency}, FAT* '19, page 220–229, New York, NY, USA. Association
  for Computing Machinery.

\bibitem[{O'Dea et~al.(2015)O'Dea, Miller, Andres, Ray, Till, and
  Saucier}]{factors_of_offensiveness}
Conor~J. O'Dea, Stuart~S. Miller, Emma~B. Andres, Madelyn~H. Ray, Derrick~F.
  Till, and Donald~A. Saucier. 2015.
\newblock \href {https://doi.org/https://doi.org/10.1016/j.langsci.2014.09.005}
  {Out of bounds: factors affecting the perceived offensiveness of racial
  slurs}.
\newblock \emph{Language Sciences}, 52:155--164.
\newblock Slurs.

\bibitem[{Penelope~Eckbert(2013)}]{language_and_gender}
Sally McConnell-Ginet Penelope~Eckbert. 2013.
\newblock \href
  {https://www.cambridge.org/de/academic/subjects/languages-linguistics/sociolinguistics/language-and-gender-2nd-edition?format=PB&isbn=9781107659360}
  {\emph{Language and Gender}}.
\newblock Cambridge University Press.

\bibitem[{Rahman(2012)}]{nword_aae}
Jacquelyn Rahman. 2012.
\newblock \href {https://doi.org/10.1177/0075424211414807} {The n word: Its
  history and use in the african american community}.
\newblock \emph{Journal of English Linguistics}, 40(2):137--171.

\bibitem[{Rudinger et~al.(2018)Rudinger, Naradowsky, Leonard, and
  Van~Durme}]{rudinger-etal-2018-gender}
Rachel Rudinger, Jason Naradowsky, Brian Leonard, and Benjamin Van~Durme. 2018.
\newblock \href {https://doi.org/10.18653/v1/N18-2002} {Gender bias in
  coreference resolution}.
\newblock In \emph{Proceedings of the 2018 Conference of the North {A}merican
  Chapter of the Association for Computational Linguistics: Human Language
  Technologies, Volume 2 (Short Papers)}, pages 8--14, New Orleans, Louisiana.
  Association for Computational Linguistics.

\bibitem[{Sap et~al.(2019)Sap, Card, Gabriel, Choi, and
  Smith}]{aae_bias_hatespeech}
Maarten Sap, Dallas Card, Saadia Gabriel, Yejin Choi, and Noah~A. Smith. 2019.
\newblock \href {https://doi.org/10.18653/v1/P19-1163} {The risk of racial bias
  in hate speech detection}.
\newblock In \emph{Proceedings of the 57th Annual Meeting of the Association
  for Computational Linguistics}, pages 1668--1678, Florence, Italy.
  Association for Computational Linguistics.

\bibitem[{Sap et~al.(2020)Sap, Gabriel, Qin, Jurafsky, Smith, and
  Choi}]{social_bias_frames}
Maarten Sap, Saadia Gabriel, Lianhui Qin, Dan Jurafsky, Noah~A. Smith, and
  Yejin Choi. 2020.
\newblock \href {https://doi.org/10.18653/v1/2020.acl-main.486} {Social bias
  frames: Reasoning about social and power implications of language}.
\newblock In \emph{Proceedings of the 58th Annual Meeting of the Association
  for Computational Linguistics}, pages 5477--5490, Online. Association for
  Computational Linguistics.

\bibitem[{Spears(2015)}]{spears2}
Arthur~K. Spears. 2015.
\newblock African american standard english.
\newblock In Green Lisa~J. Bloomquist, Jennifer and Sonja~L. Lanehart, editors,
  \emph{The Oxford Handbook of Innovation}, chapter~43, pages 786--799. Oxford
  University Press, Oxford.

\bibitem[{Spears and Hinton(2010)}]{spears1}
Arthur~K. Spears and Leanne Hinton. 2010.
\newblock \href
  {https://doi.org/https://doi.org/10.1111/j.1548-7466.2010.01065.x} {Languages
  and speakers: An introduction to african american english and native american
  languages}.
\newblock \emph{Transforming Anthropology}, 18(1):3--14.

\bibitem[{Swamy et~al.(2019)Swamy, Jamatia, and
  Gamb{\"a}ck}]{hate_speech_cross_dataset}
Steve~Durairaj Swamy, Anupam Jamatia, and Bj{\"o}rn Gamb{\"a}ck. 2019.
\newblock \href {https://doi.org/10.18653/v1/K19-1088} {Studying
  generalisability across abusive language detection datasets}.
\newblock In \emph{Proceedings of the 23rd Conference on Computational Natural
  Language Learning (CoNLL)}, pages 940--950, Hong Kong, China. Association for
  Computational Linguistics.

\bibitem[{Tatman(2017)}]{tatman-2017-gender}
Rachael Tatman. 2017.
\newblock \href {https://doi.org/10.18653/v1/W17-1606} {Gender and dialect bias
  in {Y}ou{T}ube{'}s automatic captions}.
\newblock In \emph{Proceedings of the First {ACL} Workshop on Ethics in Natural
  Language Processing}, pages 53--59, Valencia, Spain. Association for
  Computational Linguistics.

\bibitem[{Volkova et~al.(2013)Volkova, Wilson, and Yarowsky}]{volkova}
Svitlana Volkova, Theresa Wilson, and David Yarowsky. 2013.
\newblock \href {https://aclanthology.org/D13-1187} {Exploring demographic
  language variations to improve multilingual sentiment analysis in social
  media}.
\newblock In \emph{Proceedings of the 2013 Conference on Empirical Methods in
  Natural Language Processing}, pages 1815--1827, Seattle, Washington, USA.
  Association for Computational Linguistics.

\bibitem[{Waseem(2016)}]{are_you_racist}
Zeerak Waseem. 2016.
\newblock \href {https://doi.org/10.18653/v1/W16-5618} {Are you a racist or am
  {I} seeing things? annotator influence on hate speech detection on
  {T}witter}.
\newblock In \emph{Proceedings of the First Workshop on {NLP} and Computational
  Social Science}, pages 138--142, Austin, Texas. Association for Computational
  Linguistics.

\bibitem[{Wiegand et~al.(2019)Wiegand, Ruppenhofer, and
  Kleinbauer}]{hatespeech_biased_datasets}
Michael Wiegand, Josef Ruppenhofer, and Thomas Kleinbauer. 2019.
\newblock \href {https://doi.org/10.18653/v1/N19-1060} {{D}etection of
  {A}busive {L}anguage: the {P}roblem of {B}iased {D}atasets}.
\newblock In \emph{Proceedings of the 2019 Conference of the North {A}merican
  Chapter of the Association for Computational Linguistics: Human Language
  Technologies, Volume 1 (Long and Short Papers)}, pages 602--608, Minneapolis,
  Minnesota. Association for Computational Linguistics.

\bibitem[{Xia et~al.(2020)Xia, Field, and
  Tsvetkov}]{adversarial_training_aae_hatespeech}
Mengzhou Xia, Anjalie Field, and Yulia Tsvetkov. 2020.
\newblock \href {https://doi.org/10.18653/v1/2020.socialnlp-1.2} {Demoting
  racial bias in hate speech detection}.
\newblock In \emph{Proceedings of the Eighth International Workshop on Natural
  Language Processing for Social Media}, pages 7--14, Online. Association for
  Computational Linguistics.

\bibitem[{Zhou et~al.(2021)Zhou, Sap, Swayamdipta, Choi, and
  Smith}]{hatespeech_debias_words}
Xuhui Zhou, Maarten Sap, Swabha Swayamdipta, Yejin Choi, and Noah Smith. 2021.
\newblock \href {https://aclanthology.org/2021.eacl-main.274} {Challenges in
  automated debiasing for toxic language detection}.
\newblock In \emph{Proceedings of the 16th Conference of the European Chapter
  of the Association for Computational Linguistics: Main Volume}, pages
  3143--3155, Online. Association for Computational Linguistics.

\end{thebibliography}
\bibliographystyle{acl_natbib}

\appendix

\end{document}